# A Scenario-Based Mobile Application for Robot-Assisted Smart Digital Homes

Ali Reza Manashty, Amir Rajabzadeh and Zahra Forootan Jahromi
Department of Computer Engineering
Razi University
Kermanshah, Iran
a.r.manashty@gmail.com, rajabzadeh@razi.ac.ir, zahra.forootan@gmail.com

*Abstract*—Smart homes are becoming more popular, as every day a new home appliance can be digitally controlled. Smart Digital Homes are using a server to make interaction with all the possible devices in one place, on a computer or webpage. In this paper we designed and implemented a mobile application using Windows Mobile platform that can connect to the controlling server of a Smart Home and grants the access to the Smart Home devices and robots everywhere possible. UML diagrams are presented to illustrate the application design process. Robots are also considered as devices that are able to interact to other object and devices. Scenarios are defined as a set of sequential actions to help manage different tasks all in one place. The mobile application can connect to the server using GPRS mobile internet and Short Message System (SMS). Interactive home map is also designed for easier status-checking and interacting with the devices using the mobile phones.

## I. INTRODUCTION

Smart homes are becoming more popular, every day a new home appliance can be controlled digitally. New wireless technologies also help the integration of remote controls into regular mobile devices so that users can control all the different appliances using a single device. As the number of devices in the home increase, it will become harder for manufacturers to adopt a universal standard for application controls.

Ease of access and use, is the main purpose of many remote controllers we now use for our devices. Their number is getting bigger and bigger each day, as a new device becomes remotely controllable. Speakers, air conditioners, lights, curtains, garage door, TVs and players are already being remote controlled. Now that every single part of our homes can be controlled remotely, we must think of a single remote control for all of the possible actions we want to take in the house.

Smart Home is not a new term for science society but is still far more away from people's vision and audition. This is because although recent various works has been done in designing the general overview of the possible remote access approaches for controlling devices [1], or in cases simulating the Smart Home itself [2], and designing the main server [3], the design and implementation of an off-the-shelf smart home remote control application has been limited to simply the computer applications and just in cases mobile [4] and web application development [5]. Nowadays people spent a noticeable amount of time in transportation, without having access to their PCs or having hard time accessing their laptops; instead, they are constantly using their cell-phones/PDAs. Because of this, we designed and implemented a mobile application that can be connected to a server where other access routes such as web application and local windows application are also available in there.

An important question regarding the problem is whether developing the web application can take the place of mobile phone applications, due to availability of web sites through GPRS and WiMax wireless internet? The answer to this question is that even though we can access our home control system through mobile wireless internet and use of current mobile browsers, which are now no less powerful than PC browsers, they cannot access GSM messaging systems such as SMS, MMS and so on. On the other hand, simultaneous accessing to mobile internet services (e.g., GPRS) for loading a complete webpage, is still expensive, so there will not be any need for designing the home control schedules and rules [5] online; instead, a temporary connection will do the information updating while using the mobile application offline.

In this paper, we present design of a mobile application for accessing and controlling the smart home control systems. We also show an implementation overview using Windows Mobile platforms and C# language as well as the general outline of the system. The term scenario is defined in this paper and the way of specifying robot-assisted tasks is also described.

## II. SYSTEM ARCHITECTURE

### A. Preface

The Smart Home system usually consists of several devices scattered around the house that are linked together using a wired or wireless network. A computer system acts as the server in a node which controls all the information exchange through the network. The server system must have a device manager -middleware- [3] that assists the main application by connecting all the different device controllers through a single interface or the least interfaces possible.

### B. Devices

The various types of devices in this house can be divided into three categories:

1. Actuator devices, e.g., alarms, lights and doors

2. Sensor devices, e.g., heat, movement and healthcare

3. Actuator/Sensor devices, e.g., robots, air conditioners

All the devices (whether sensor or actuator) can express their status using their controllers, such as whether they are on or off or the job they are currently involving (e.g. closing the door).

A device driver might also be needed for more complex devices; because each device might be anything from a lamp to a home robot.

### C. Robots

Robots can be regarded as an actuator/sensor device, yet with wider range of abilities. Robots can be used in future homes more efficiently than the devices. This can be done by using robots for many actions that different devices can do independently. Robots (like Joy steward robot [6]) have many different capabilities that can be used in a smart home control system. We can use robots in many places where devices cannot be controlled independently, like moving and cleaning other device. Using the robots can eventually lower the number of controllable devices to as low as the number of robots in a home.

Another advantage of using robots instead of controllable devices is that there will not be any need for connecting different devices in predefined positions in the home; instead robots can move to any place in the home, controlling and monitoring all the possible devices a home can contain. This will help integrating the home with a smart robot far faster and easier than installing the controller for different devices. On many devices, installing controllers may also be difficult or even impossible.

Because current commercial robots, like cleaning bots, are able to do a limited set of actions a combination of controllable devices and robots is recommended. So for controlling devices using robots, we must make a three level scheduling schema in which, first the robot, then the device and then the corresponding device action must be selected.

### D. Subsystems

The server application has the capability of adding newly installed devices and providing the appropriate controlling methods. The controlling signal and status controlling schema for all these devices is available through a general interface. We refer any of these devices, as objects. So the application must be object-independent while the device manager is closely in contact with all these objects through their device drivers and appropriate connections (e.g., cable or wireless Ethernet) as King et al. designed Atlas platform as a middleware in this field [7]. Using the object-independent programming interface, we can extend the controlling methods to any further possible ways such in the web application, mobile application and telephone line controller; easily without the need of changing the application codes. The subsystems in this smart home controlling environment is illustrated is Fig. 1.

The database of the system is playing an important role. The connection of the web application and mobile application to the home control system will be through the information in the database.

There are several parts of the system that must be designed and linked together. From the subsystems showed in Fig. 1, we are going to design and implement the mobile application, because, as mentioned in the Introduction section, the other systems had been designed before.

Before designing the application itself, we must design the server connection schema of the system, in which the connections of the different parts are modeled.

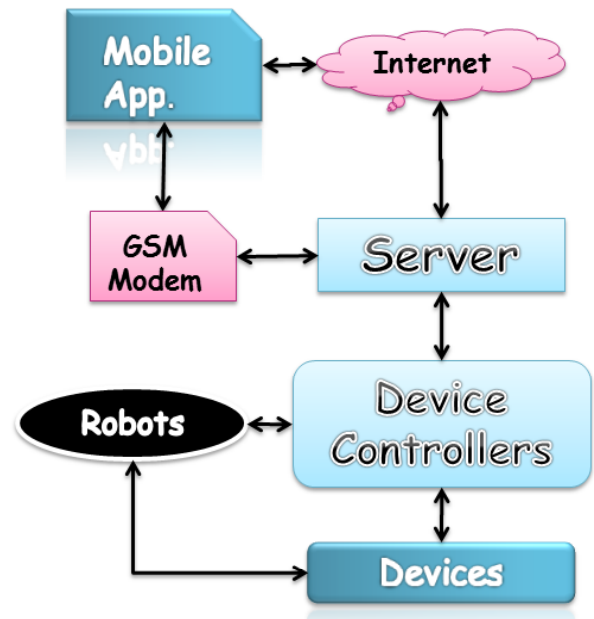

Figure 1. The subsystems overview of a robot-assisted Smart Home controlling system using a mobile application.

### E. Server

The server is actually the computer system in the smart home that contains the windows application and device

(IJCSIS) International Journal of Computer Science and Information Security,
Vol. 8, No. 5, August 2010manager. The server retrieves the sensor information on regular intervals and updates the database. This interval is different for vital and non-vital devices. For example, elderly and disabled person's health monitoring sensors [8, 9] data must be updated at least every second. The burglary detection system can be updated every few seconds but the temperature and light sensors might be checked within minutes. These categories help energy-saving schemas to be applicable.

This database also contains information about the devices, scenarios, rules, user access and other policies. Other remote control applications, such as web and mobile application, retrieve sensor and device information from this database and update the scenarios and rules accordingly so each time server updates the database sensor information, it also checks the changes applied to rules and scenarios and perform the actions necessary.

The web application can connect directly to the database, but because of functional restriction, mobile devices cannot connect directly to the server database and update the information like regular data connections. So here we face two important issues. The first issue is where to keep the database, so remote application can have continuous access to it? The answer to this question is completely related to the web application and security. Web applications can be either hosted in the Smart Home server or a host and domain reseller server. Due to security measures, we decided to make the web server and the database all in same place in the home server. This requires a static IP for the house, which is not a problem as security comes first. So the location of the web application is in the Smart Home server and the database is shared between the web and windows application. The second issue is the connection of the mobile application to database for retrieving and updating the information. As mentioned before, current mobile applications may not have the required memory and libraries required to establishing a direct connection to the server, so the only way we can exchange information is through http web servers using GPRS (small data packets can be sent through SMS to the GSM modem attached to server). One possible way to do this task is using web requests. Web requests are parameters send to web site using the "?" operator after the webpage name (e.g., www.test.test/login.aspx?user=admin&&pass=123456).
After processing the web page requests, it can be identified that a request has been set using the mobile application (using the appropriate web *requests*), and the page *response* will be changed according to the *request*.

The web response page is regularly the html content of the website, but using ASP.Net application, before the page can be loaded, according to the web *requests* we can send limited lines of information instead of the whole html page. So we can use this feature to exchange information between the mobile application and a simple aspx page we already developed in our web server.

Now we described the overall server behavior. From here on, we focus on the mobile application design and implementation.

III. SCENARIO-BASED MOBILE APPLICATION DESIGN

*A. Scenarios*

Scenarios are also very impressive in this design. Scenarios are list of actions that consists of other different scheduled tasks or scenarios. A scheduled task is simply a scenario with only a single task.

These scenarios make it easy for people saving the list of actions for further use, in addition to design multiple actions to be done in a single scenario. Later the scenarios can be enabled/disabled in the scenarios list or be used in another scenario too. Cheng, Wang and Chen proposed a reasoning system for smart homes that is also scenario based [10].

Here with robots able to do different actions to other devices, it may be necessary also to select the robot in first place and the device and then the action the robot must do with the device. Since different robots can do different things on different devices, when the desired robot is selected, the appropriate devices and actions are available for the robot. For example a robot can move things around the house, the other one can clean different things, one may cook and another one will check the different rooms of the house for intruders and phenomena (e.g., catching fire and breaking of water pipes).

An example of a scenario of the scheduled tasks is as below:

Scenario name: Watering Plants

A. Sprinkler 1: on @ 5:00 AM

B. Sprinkler 2: on @ 5:30 AM

C. Sprinkler 1: off @ 7:00 AM

D. Sprinkler 2: off @ 9:00 AM

An important fact we must consider is that although many actions on devices are single state tasks, another tasks might need some parameters. For example when we want to set the temperature of an air conditioner system, we might also need to pass the desired temperature to the device. This is more important and complex when using special robots. For example when a robot is capable of moving or cleaning the objects, we must use a term for the action that both indicate the action and the subject of the action. We show parameters of an action in parentheses.

Only one time a user sets the actions in the scenario and then he or she can use it several times after defining it. Another advantage of using scenarios is that we can use other scenarios in the scenario we are currently defining. Example of scenarios involving robots and other scenarios is as below:

Scenario name: Clean Home

A. Cleaning robot: Clean (Bathtub) @ Now

B. [Gather Dishes] @ 10:00 AM

C. Home robot→Washing machine: on @ 10:05 AM

D. Cleaning robot: Clean (Saloon) @ 10:05 AM

The above *Clean Home* scenario consists of three scheduled tasks and one scenario called *Gather Dishes*. Scheduled time of the scenarios will be overridden when a new time is set (like task *B* in above scenario). Scenarios used in another scenario are placed in brackets ([ ]). *Washing machine* and *Cleaning robot* are considered as devices here. The words followed by them after ":" are the actions they must take. When a robot is used to do an action of a device, the first item of the scenario action will be the name of the robot doing the action followed by an arrow (→) (e.g., Home robot will turn the Washing machine on, not the machine itself). In this scenario, *Clean* action of the *Cleaning robot* is parameterized so that the robot knows how to come along with the parameter.

Another example is the *Gather Dishes* scenario that was already used in Clean Home scenario and contains parameterized actions of the *Mover robot*:

Scenario name: Gather Dishes

A. Mover robot: GoTo (Saloon) @ Now

B. Mover robot: PickUp (Dishes) @ In 2 Minutes

C. Mover robot: GoTo (Kitchen) @ In 5 Minutes

D. Mover robot: PutInto (WashingMachine) @ In 6 Minutes

E. Mover robot: GoTo (DefaultPosition) @ In 7 Minutes

*B. Use-Cases*

The main server computer, which is located in the smart home area, is loaded with the windows application that gives the administrator user a comprehensive set of options and capabilities. The user can add and manage devices and robots in the application (of course if hardware procedures had been proceeded previously), design the home top view plane using graphical tools and icons, manage user access controls (e.g., define access limits for children), define policies of remote access (e.g., authenticated phone numbers), define rules (conditions to be checked and actions taken if the predefined criteria is met), define scenarios and check the current status of devices and robots (Fig. 2).

In the other hand, mobile application user can do the most common and important tasks, but not all the functionalities, as illustrated in Fig. 2. This limitation is mainly because of limits of a mobile application implementation and security factors. For example the mobile application user cannot design the home top view plane that graphically shows the current status of the home, but can simply view it from his/her cellphone.

Mobile applications can use advantages of cellphones such as built-in microphone and color display that usual remote controls suffer from having it. This will make mobile applications capable of live streaming of cameras within the house and implementing speech recognition for ease of access [11].

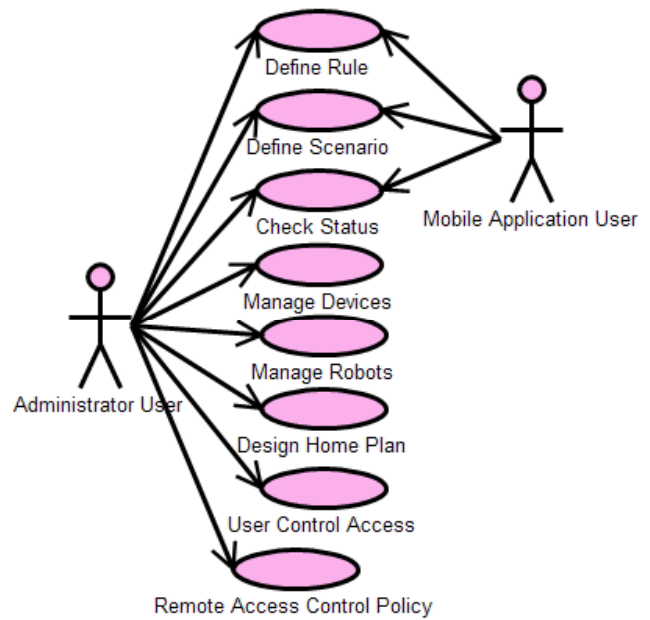

Figure 2. Use-cases of the Administrator and the Mobile user

*C. Connections to Server*

As mentioned earlier, we must connect to the database using a web server. The web application can have most of the capabilities and access levels of the windows application of the server but with some restricted regulations. We can allocate some pages of the web application for allowing access from the mobile application that responds to the web requests sent from the mobile http connection.

Because this approach of sending/receiving information is not encrypted, we use encryption algorithms known for both the web and mobile application. For extra security, we use a magic number used for hashing information that expires soon and need to be reconfigured by the web server.

The first attempt to connect to web server with the special code will gives the encrypted magic number to the mobile application. Then the user name and password will be sent using a hashing algorithm by the magic number as the salt, as the web *request* parameter. Then the server returns the authentication acknowledge back to the mobile application. Now every request from the server, such as request for updating and checking the status of the devices, must be accompanied by the hashed username and password. After some predefined time (e.g., 5 minutes), the magic number expires and the server data packets must include newly hashed username and password using the new magic number.

The above sequence is shown in Fig. 3, which also shows the sequence of main server application, retrieving information from the devices (here a robot) and database

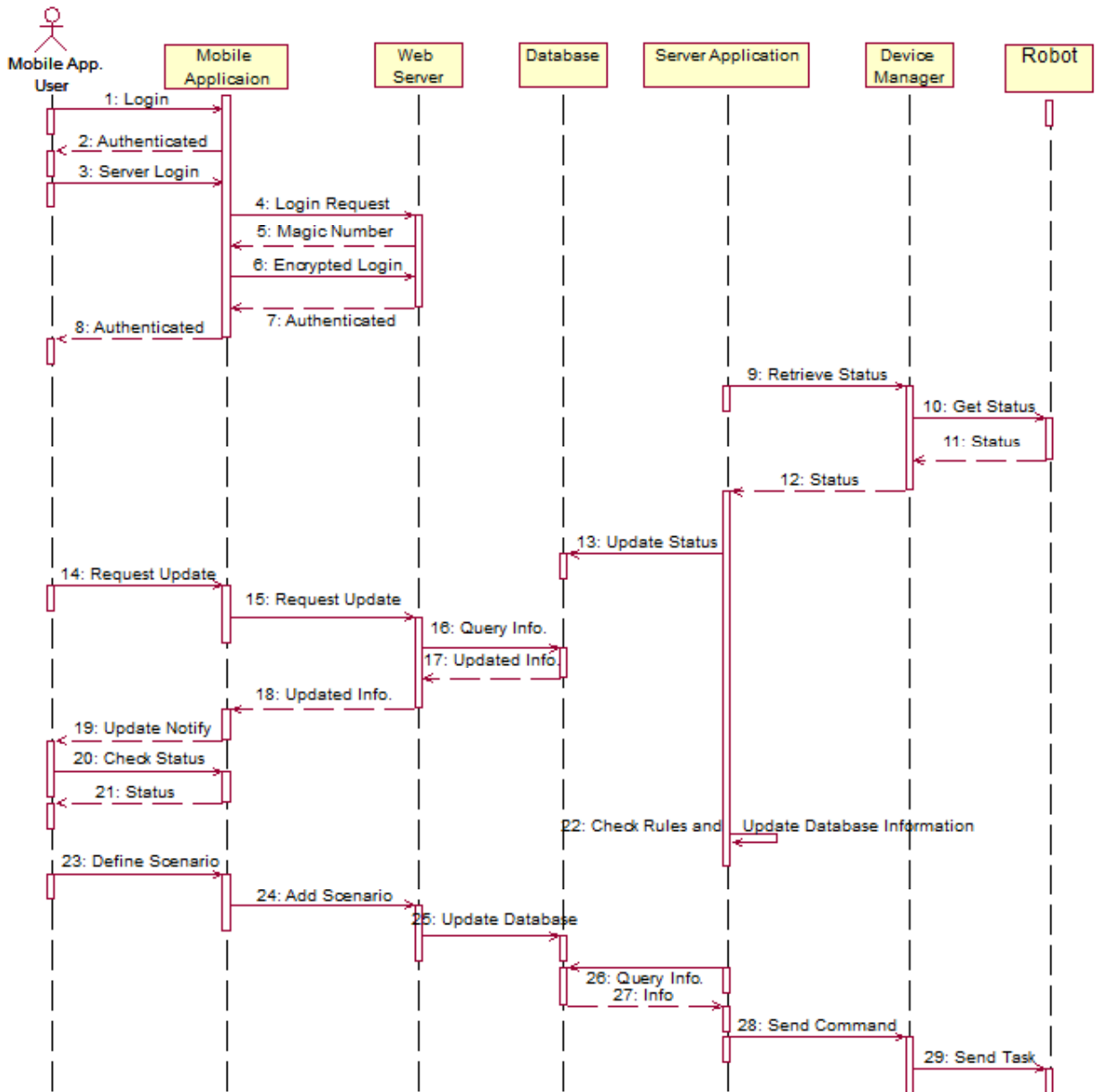

Figure 3. The sequence diagram of defining a robot-related scenario using the mobile application. This UML sequence diagram shows the necessary steps from logging into mobile application, defining the tasks and taking the appropriate actions to the robot as soon as application server becomes aware of it while making the changes necessary. As soon as the mobile application or web application updates the database, the server will check the updated information, time and conditions, so that it will send necessary commands to the actuators to make the changes applied.

The database contains a table in which icons of the devices and their respective positions will be stored. This information is also transferred to the device in the case of updating the information. This table information is related but not depended on the main device table (in which detailed device controlling information is stored). In other way, the records and items in the home top view plane can be in the device records list either, but not necessarily all the map items must have full identification record in the device records table. This makes the whole house map items easy designing, but not limiting the selection to controllable devices. Now we must find a suitable way to transfer this map to the mobile device. The problem is the file size of the map that is too much to be transferred easily through the GPRS mobile internet; instead, we designed a way to create the map in the mobile device itself, using pre-defined icons.

IV.  THE MOBILE APPLCATION IMPLEMENTATION

*A. Home Map*

A home map is necessary to simplify the access and status checking of the home devices and robots. Lu and Fu proposed an activity map for a convenient and user-accessible interaction with the smart environment [12].

To transfer the data and create the map in the mobile device, first we must separate the map plane and objects within it. To do the task, as the device requests for updating the information of the devices and the maps, we send the house wall information as arrays of connected lines. To make this thing happen, we actually need a List<List<Points>> (a List of List of Points). The inner list contains points that makes continues lines within the points of the current list, making an open polygon designing possible. The first two points of each inner list will be used to determine the width and color of the lines. Because every point has two integer elements (x and y), two points make 4 integer data integration possible, so one is used for the line width and the other three represents an RGB color value.

For transferring the icons data, we used a record with the fields illustrated in Fig. 4. *OID* field, if not zero, can represent a device in the according device table in the database that can make the device selectable (e.g., for further details view and scenario/rule assignment). Other devices in the map have an *OID* of zero. *Name* and *Position* fields of the device map record are used for displaying purposes. The *IconID* field, regardless of the status of the device in the appropriate device table record, indicates the current status of the device/furniture using an icon in the mobile application icon database. The web page map controller is responsible for representing the appropriate *IconID* that best defines the *type* and *current status* of the device. For example, two icons can represent a door in two different statuses of being closed or opened.

| OID | Name | Position | IconID |
| --- | --- | --- | --- |

Figure 4. Fields in the record used to transfer home top view plane icons to the mobile application

For new devices that their icons is not available in the mobile application, some extra icons has been considered that makes the other unknown devices into 4 categories that can be recognized by their status easily:

1. On/Off devices (e.g., a lamp)

2. Leveled Devices (e.g., a gas sensor)

3. Appearing/Disappearing devices (e.g., a car or a bike)

4. Opened/Closed Devices (e.g., a door)

The device then repaints the map using the received information from the server, and the icons in its database. First the areas of the house are drawn using the line information defined by the points and the icons are painted just after it. The controllable devices in the map (whose OID is not zero) can be pointed and selected (like the application of the Gator Tech Smart House [3]) to check the status and define scenarios for it.

### B. Retrieving Information from Server

For implementing the updating procedure, we divide the updating data into two categories. The first part is the devices data table which includes information about the devices, as well as the capabilities and controllable parts of each one. Because this information may be quite large to transfer and the devices and their controllable/sense-able features is device dependent, but not state dependent; they can be updated in longer periods than state information. This type of updating is labeled *Update Devices Data* in the main menu of the mobile application

The second part of the information is the device states and map information. Because this information is more likely to be updated, and contain less data than the first part, they can be downloaded every time the statuses are being checked. This regular information updating is accessible as *Update Information* button in the main menu form of the mobile application, as well as in *Check Status* and *Home Top Plane View* forms.

### C. Windows Mobile platform Implementation

There are several platforms in which it is possible to implement the designed application, such as Java 2 Micro Edition (J2ME), Windows Mobile and Symbian. The J2ME is the most common platforms supported in mobile devices, but it's low level libraries makes it difficult to implement the application in the first place after designing it. So we decided to implement the application, in this stage, using Windows Mobile platform, which can help implementing the application with all the possible features as necessary.

*1) Login Screens*

For increased security, both the mobile application and server connection require a username and password. This was done so that the server address can be also protected from unauthorized viewing.

  *a) User Login*

This screen simply contains the username and password fields for the user to access the application.

  *b) Server Login*

Just like the simple User Login form, but with an additional textbox field labeled *Server path* that indicates the home web server address that the mobile application must connect to communicate with the server application.

*2) Main Menu items*

The main menu form of the application is the form appearing just after logging in. This form contains all the links necessary for different parts of the program (Fig. 5).

*Update Information* and *Update Devices data* was mentioned before. The *Home Top Plane View* will bring up the home top plane map of the house using the most recent updated information.

*Live Camera Streaming* is designed for live streaming the camera devices in the smart home. The *Manage Scenarios* and *Manage Rules* items are designed to list and manage the current device scenarios and conditional rules. Scenarios are set of tasks which will be done in a specific time (or current moment). Rules are condition and action sets in which the

*(IJCSIS) International Journal of Computer Science and Information Security,*
*Vol. 8, No. 5, August 2010*

condition are simultaneously checked and as soon as the criteria is met; the appropriate actions will be taken.

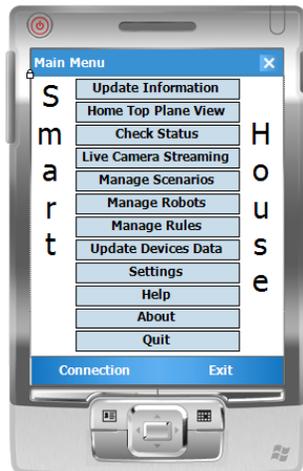

Figure 5. The main menu form of the mobile application. All different parts of the application can be accessed through this form.

*Manage Robots* will provide options regarding the robots, such as the list of all of them and the specific action each robot can do. The list is read-only in mobile application but the robots or their actions can be enabled/disabled for selecting in the scenario tasks.

The settings option will allow the user to change the server settings such as the address and update rules; as well as changing username/password and other settings.

*3) Home Map*

This form shows the home top plane map of the house, designed by the administrative user in the server computer, and downloaded as records of map elements by the mobile application (Fig. 6). If the information is out of date, it prompts the user to update the information. The user can also select the devices here and be prompted whether he/she wants to check its status or add a scenario for it and then will be redirected to the appropriate form. The icons in this map also represent the current status of the device as indicated by the *IconID* field of the device map record retrieved from the server.

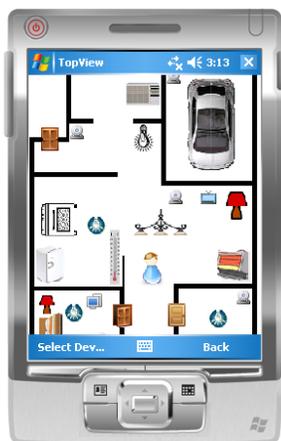

Figure 6. Home top view plane of the house with selectable items.

*4) Managing Scenarios*

In this form, the updated scenarios will be listed and each of them can be modified and be enabled/disabled. New scenarios can also be created in a new form (Fig. 7). Each task has a name and will be applied on a device by the corresponding device action. The time of the scenario can be set to now, a specific time in the day or a specific time after execution time. Other defined scenarios can also be selected as a task in the new scenario. When a new task is selected to be designed, different robots can also be selected as the actor of the action and each action can take a value as the parameter. These data can be sent using SMS either.

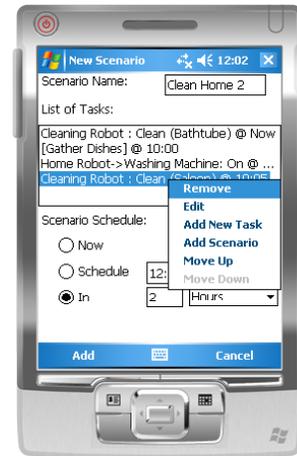

Figure 7. The New Scenario form of the mobile application. A list of tasks can be defined here including other scenarios or robot actions.

## V. FUTURE WORKS

As mentioned earlier, the works on a complete and comprehensive smart home that can work with all possible home appliances and can be controlled by all means in an effective way are all scattered around and researched independently. Only in some projects, some parts of a real smart home put into practice (e.g., The Gator Tech Smart House [13], and Plug and Play smart environment [14]). As we completed a comprehensive design from the previous work [15] and designed a server and mobile application needed for controlling the smart home remotely, we must continue the work on completing the whole server and applications, and move through commercial manufacturing of such houses so that all these efforts on designing the Smart Home can come to reality.

## VI. CONCLUSION

In this paper, we presented an overview of a smart home control system server along with the way the devices are managed in the server. After discussing the possible security issues of developing a server for communicating to mobile application, we proposed a web server for the mobile application to communicate to it using GPRS. We presented the communication sequence through the web server in a UML sequence diagram and described the use-cases of both the server and mobile application. The scenarios were designed to set a number of tasks all in one place for further and easier use. Robots participation and parameterized actions were also described along with regular actions. We finally explained the design of the mobile application and the data records needed for transferring the data and home top view plane from the server to mobile application in an efficient way. We finally described the main parts of the implementation of this smart home remote control mobile application in the Windows Mobile platform.

## REFERENCES

[1] A. Sleman, M. Alafandi, and R. Moeller,"Integration of Wireless Fieldbus and Wired Fieldbus for Health Monitoring," Consumer Electronics, 2009. ICCE '09. Digest of Technical Papers International Conference, pp.1 – 2, 10-14 Jan. 2009

[2] T. Van Nguyen, J. Gook Kim, and D. Choi, "ISS: The Interactive Smart home Simulator," Advanced Communication Technology, 2009. ICACT 2009. 11th International Conference on, vol. 03, pp.1828-1833, 15-18 Feb. 2009

[3] C. Escoffier, J. Bourcier, P. Lalanda, and Y. Jianqi, "Towards a Home Application Server," Consumer Communications and Networking Conference, 2008. CCNC 2008. 5th IEEE, pp.321-325, 10-12 Jan. 2008

[4] Z. Salvador; R. Jimeno, A. Lafuente, M. Larrea, and J. Abascal, "Architecture for ubiquitous enviroments" Wireless And Mobile Computing, Networking And Communications, 2005. (WiMob'2005), IEEE International Conference on, pp.90-97, vol. 4, 22-24 Aug. 2005

[5] G. Patricio, and L. Gomes ,"Smart house monitoring and actuating system development using automatic code generation," Industrial Informatics, 2009. INDIN 2009. 7th IEEE International Conference on, pp.256-261, 23-26 June 2009

[6] H. Lee, Y Kim, K. Park, Z. Zenn Bien, "Development of a steward robot for human-friendly interaction," Computer Aided Control System Design, 2006 IEEE International Conference on Control Applications, 2006 IEEE International Symposium on Intelligent Control 2006 IEEE, pp.551 – 556, 4-6 Oct. 2006

[7] J. King, R. Bose, Y. Hen-I, S. Pickles, A. Helal,"Atlas: A Service-Oriented Sensor Platform: Hardware and Middleware to Enable Programmable Pervasive Spaces," Local Computer Networks, Proceedings 2006 31st IEEE Conference on, pp. 630 – 638, 14-16 Nov. 2006

[8] V. Santos, P. Bartolomeu, J. Fonseca, and A. Mota,"B-Live - A Home Automation System for Disabled and Elderly People," Industrial Embedded Systems, 2007. SIES '07. International Symposium on, pp.333-336, 4-6 July 2007

[9] H. Medjahed, D. Istrate, J. Boudy, and B. Dorizzi, "Human activities of daily living recognition using fuzzy logic for elderly home monitoring," Fuzzy Systems, 2009. FUZZ-IEEE 2009. IEEE International Conference on, pp.2001-2006, 20-24 Aug. 2009

[10] Yerrapragada, C.; Fisher, P.S.;" Voice Controlled Smart House"; Consumer Electronics, 1993. Digest of Technical Papers. ICCE., IEEE 1993 International Conference on, pp.154-155, 8-10 June 1993

[11] Sheng-Tzong Cheng; Chi-Hsuan Wang; Ching-Chung Chen; "An Adaptive Scenario Based Reasoning system cross smart houses" Communications and Information Technology, 2009. ISCIT 2009. 9th International Symposium on, pp.549 – 554, 28-30 Sept. 2009

[12] C. Lu, L. Fu, "Robust Location-Aware Activity Recognition Using Wireless Sensor Network in an Attentive Home," Automation Science and Engineering, IEEE Transactions on, vol. 6, Issue.4, pp.598-609, Oct. 2009

[13] S. Helal, W. Mann, H. El-Zabadani, J. King, Y. Kaddoura, and E. Jansen, "The Gator Tech Smart House: a programmable pervasive space", Computer, vol. 38, Issue. 3, pp.50 – 60, March 2005

[14] B. Abdulrazak, A. Helal, "Enabling a Plug-and-play integration of smart environments", Information and Communication Technologies, 2006. ICTTA '06. 2nd vol. 1, pp.820 – 825, 2006

[15] A. Rajabzadeh, A. R. Manashty, and Z. Forootan Jahromi, "A Mobile Application for Smart House Remote Control System," International Conference on Wireless Communication and Mobile Computing (ICWCMC 2010), Proceedings of WASET, vol. 62, ISSN. 2070-3724, pp. 80-86, February 2010

AUTHORS PROFILE

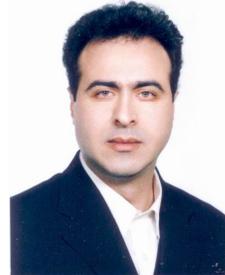

**Amir Rajabzadeh** received the B.S. degree in telecommunication engineering from Tehran University, Iran, in 1990 and received the M.S. and Ph.D. degrees in computer engineering from Sharif University of Technology, Iran, in 1999 and 2005, respectively. He is currently an assistant professor of Computer Engineering at Razi University, Kermanshah-Iran. He was the Head of Computer Engineering Department (2005–2008) and the Education and Research Director of Engineering Faculty (2008-2010) at Razi University.

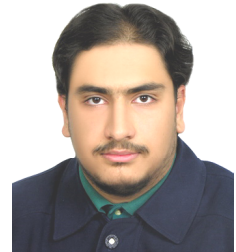

**Ali Reza Manashty** is a senior B.S. student in Software Computer Engineering at Razi University, Kermanshah, Iran and will be graduated in September 2010 awarding the 3rd rank among the autumn 2006 entrance students of Computer Engineering Department. He is going to continue his academic career as a M.Sc. student in the next semester. He has been researching on mobile application design and smart environments especially smart digital houses since 2009. His publications include 4 papers in international journals and conferences and one national conference paper. He has earned several national and international awards regarding mobile applications developed by him or under his supervision and registered 4 national patents. He is a member of Exceptional Talented Students office of Razi University since 2008 and he was the teacher assistant of several under-graduate courses since 2008.

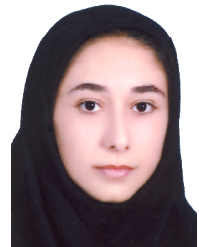

**Zahra Forootan Jahromi** is a senior B.S. student in Software Computer Engineering at Razi University, Kermanshah, Iran and will be graduated in September 2010. She is now researching in smart environments specially on simulating smart digital homes. Her publications include 4 papers in international journals and conferences and one national conference paper. She has 3 registered national patents and is now teaching Robocop robot designing for elementary and high school students at Alvand guidance school. She is a member of Exceptional Talented Students office of Razi University since 2008.